\documentclass{article} 
\usepackage{iclr2026_conference,times}
    
\usepackage{amsmath,amsfonts,bm}

\def\eqref#1{equation~\ref{#1}}

\def\1{\bm{1}}

\DeclareMathAlphabet{\mathsfit}{\encodingdefault}{\sfdefault}{m}{sl}
\SetMathAlphabet{\mathsfit}{bold}{\encodingdefault}{\sfdefault}{bx}{n}

\usepackage{hyperref}
\usepackage{url}

\usepackage{graphicx} 
\usepackage{enumitem}
\setlist[enumerate]{nosep}

\usepackage{xcolor}

\usepackage[utf8]{inputenc} 
\usepackage{booktabs}       
\usepackage{amsfonts}       
\usepackage{nicefrac}       
\usepackage{microtype}      
\usepackage{lipsum}         
\usepackage{natbib}
\usepackage{doi}
\usepackage{verbatim}
\usepackage{multirow}
\usepackage{amsmath}
\usepackage[linesnumbered, ruled, vlined]{algorithm2e}
\usepackage{listings}
\usepackage{amssymb}   
\usepackage{CJKutf8}
\usepackage{chngcntr}
\usepackage{cleveref}       
\usepackage{fancyhdr}
\usepackage{makecell}

\SetKwInput{Input}{Input}
\SetKwInput{Output{Output}}

\title{CRAFT-GUI: Curriculum-Reinforced Agent For GUI Tasks}
    
\author{
  Songqin Nong\textsuperscript{1}, Xiaoxuan Tang\textsuperscript{1}, Jingxuan Xu\textsuperscript{1}, Sheng Zhou\textsuperscript{2}, \\
  \textbf{Jianfeng Chen\textsuperscript{1}, Tao Jiang\textsuperscript{1}, Wenhao Xu\textsuperscript{1,}\thanks{Corresponding Author}} \\
  \textsuperscript{1}Ant Group, \textsuperscript{2}Zhejiang University \\
  \texttt{\{nongsongqin.nsq, tangxiaoxuan.txx, xujingxuan.xjx,} \\
  \texttt{jianfengchen.cjf, tara.jt, hao.xuwh\}@antgroup.com} \\
  \texttt{zhousheng@zju.edu.cn}
}

\iclrfinalcopy 
\begin{document}

\maketitle

\begin{abstract}
As autonomous agents become adept at understanding and interacting with graphical user interface (GUI) environments, a new era of automated task execution is emerging.
Recent studies have demonstrated that Reinforcement Learning (RL) can effectively enhance agents' performance in dynamic interactive GUI environments.

However, these methods face two key limitations: (1) they overlook the significant variation in difficulty across different GUI tasks by treating the entire training data as a uniform set, which hampers the agent’s ability to adapt its learning process; and (2) most approaches collapse task-specific nuances into a single, coarse reward, leaving the agent with a uniform signal that yields inefficient policy updates.
To address these limitations, we propose \textbf{CRAFT-GUI}, a curriculum learning framework based on Group Relative Policy Optimization (GRPO) that explicitly accounts for the varying difficulty across trajectories. To enable more fine-grained policy optimization, we design a reward function that combines simple rule-based signals with model-judged evaluation, providing richer and more nuanced feedback during training. 
Experimental results demonstrate that our method achieves significant improvements over previous state-of-the-art approaches, outperforming them by 7.1\% on public benchmarks AndroidWorld and 10.3\% on our private constructed dataset, respectively. These findings empirically validate the effectiveness of integrating reinforcement learning with curriculum learning in GUI interaction tasks.
\end{abstract}

\section{Introduction}

Graphical User Interface (GUI) agents offer a promising approach for automating software interactions by autonomously perceiving and manipulating visual elements. Unlike traditional automation tools~\cite{rawles2023androidinthewild, wang2025mobile} that depend on structured inputs, modern GUI agents~\cite{qin2025ui, gou2024navigating, lin2025showui} leverage multimodal large language models to directly interpret GUI screenshots and perform user-specified tasks, enabling robust automation even without API access.

\begin{figure}[t]
    \centering
    \includegraphics[width=\linewidth, height=0.4\textheight, keepaspectratio]{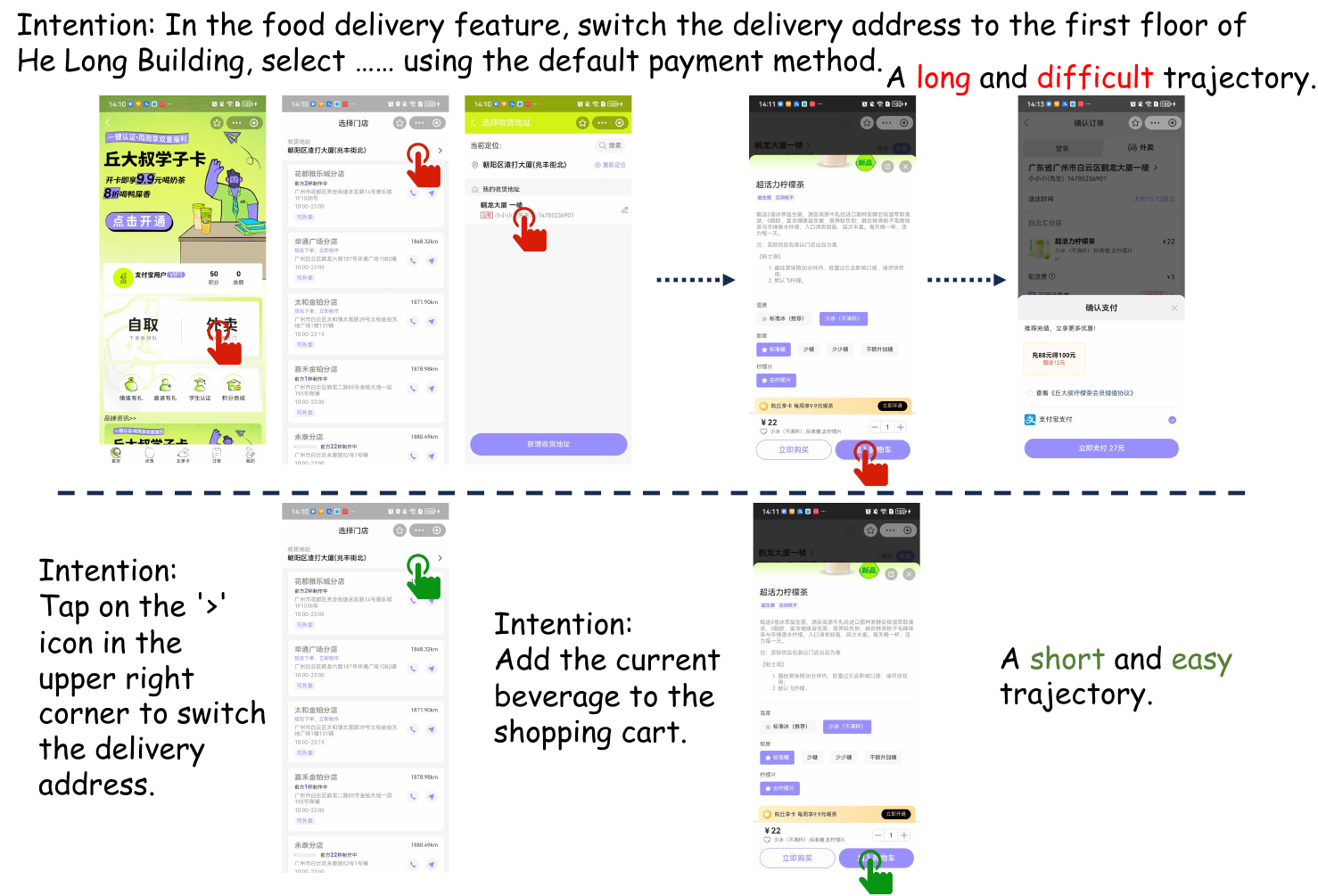}
    \caption{Contrast between complex and simple GUI tasks.}
    \label{fig:fig1_data}
    \vspace{-1.5em}
\end{figure}

Although recent studies have made progress, current GUI agents~\cite{cheng2024seeclick, gou2024navigating, lu2025ui, luo2025gui, nong2024mobileflow} typically treat all training instances uniformly. However, real-world GUI tasks exhibit substantial variation in difficulty, as shown in Figure~\ref{fig:fig1_data}. Ignoring these differences leads to two critical drawbacks: \textit{(i) optimization instability during training and (ii) constrained model capability growth}. Uniform treatment of samples impedes the agent's ability to adaptively adjust its learning based on task complexity—analogous to expecting a child to simultaneously master elementary and university curricula.

Furthermore, when applying reinforcement learning with verified rewards, existing reward functions rely on coarse, rule-based heuristics that fail to differentiate task difficulty. Rather than fundamentally addressing how task difficulty impacts learning, current methods~\cite{lu2025arpo} resort to intricate task selection and experience replay strategies, adding unnecessary complexity while neglecting nuanced, task-aware feedback for efficient policy optimization.

Inspired by how humans naturally progress from simple to complex GUI interactions, we propose a curriculum-aware reinforcement learning framework tailored to GUI agent training. While previous works~\cite{team2025kimi, xie2025logic, coreteam2025mimovltechnicalreport} have shown benefits of staged learning in language domains, GUI automation presents unique challenges: heterogeneity in semantic complexity, grounding requirements, and interface dependencies render simple curriculum staging insufficient. This motivates our research question: \textit{Can a difficulty-aware reinforcement learning strategy boost GUI task performance?}

Our method organizes training samples based on trajectory complexity, progressing from shorter, simpler tasks to longer, more complex ones. We incorporate fine-grained reward mechanisms providing informative feedback for tool usage, spatial precision, and semantic correctness. By integrating both operation and understanding tasks, the agent jointly develops low-level action competence and high-level task comprehension. Experimental results demonstrate significant improvements: 7.1\% gain on AndroidWorld and 10.3\% on our private benchmark over prior state-of-the-art approaches.

\vspace{0.5em}
\noindent\textbf{Contributions.}
(1) A curriculum RL strategy that systematically progresses from simple to complex tasks based on trajectory characteristics.
(2) Fine-grained hybrid reward mechanisms integrating rule-based and model-predicted evaluation for stable convergence.
(3) Comprehensive evaluation demonstrating strong improvements over SFT and RL baselines on both operation and understanding tasks.
\vspace{-0.5em}

\section{Related Work}

\subsection{GUI Agent}

Current GUI agent implementations follow two main paradigms. The first relies on closed-source multimodal models\cite{anthropic2024cuda, openai2024gpt4o} with prompt engineering, using visual input and system APIs for perception and decision-making. Representative works include AppAgent~\cite{zhang2025appagent, li2024appagent, jiang2025appagentx} and Mobile-Agent series~\cite{wang2024mobile,wang2024mobilev2}. However, these approaches face key limitations: reliance on manual prompt tuning, performance bottlenecks from general-purpose models, and limited adaptability to diverse scenarios.

The second approach trains specialized models for GUI tasks, emphasizing task-specific optimization. Notable implementations include CogAgent~\cite{hong2024cogagent}, AutoGLM~\cite{liu2024autoglm}, Ferret-UI series~\cite{you2024ferret, li2024ferret}, UI-Venus ~\cite{gu2025uivenustechnicalreportbuilding} and UI-TARS~\cite{qin2025ui}. While these methods demonstrate improved performance on specific benchmarks, they remain constrained by dependence on large-scale pretraining data and performance degradation in out-of-distribution scenarios.

\subsection{Reinforcement Learning for GUI Agents}

The RL paradigm with verifiable rewards has emerged as a promising approach, inspired by reasoning models like OpenAI o1~\cite{jaech2024openai} and DeepSeek-R1~\cite{guo2025deepseek}. In GUI domains where verifiable reward signals can be derived through predefined rules, RL-based training strategies are gaining increasing adoption. UI-R1~\cite{lu2025ui}, GUI-G2 ~\cite{tang2025guig2gaussianrewardmodeling} and GUI-R1~\cite{luo2025gui} apply rule-based GRPO training for operation tasks; InfiGUI-R1~\cite{liu2025infigui} implements SFT cold-start prior to GRPO; UIShift~\cite{gao2025uishift} restricts GRPO training to the LLM component; ARPO~\cite{lu2025arpo} proposes experience replay-optimized algorithms.

However, current implementations exhibit three fundamental limitations: (1) \textit{Data distribution bias}: none address the detrimental impact of imbalanced difficulty distributions through systematic curriculum approaches; (2) \textit{Task scope restriction}: excessive focus on operation tasks with inadequate support for understanding tasks; (3) \textit{Reward granularity}: overly simplistic reward formulations lacking multi-dimensional feedback. Our work addresses these gaps through curriculum-based RL with fine-grained, task-aware reward mechanisms.

\section{Method}
In this section, we present the key components of our method. We first revisit Group Relative Policy Optimization (GRPO)~\cite{shao2024deepseekmath} and describe how it is combined with curriculum learning to improve training stability and sample efficiency in complex GUI environments. We then introduce our task-specific reward design, followed by a detailed formulation of the training procedure.

\begin{figure}[!t]  
    \centering  
    \includegraphics[width=\textwidth, height=0.25\textheight]{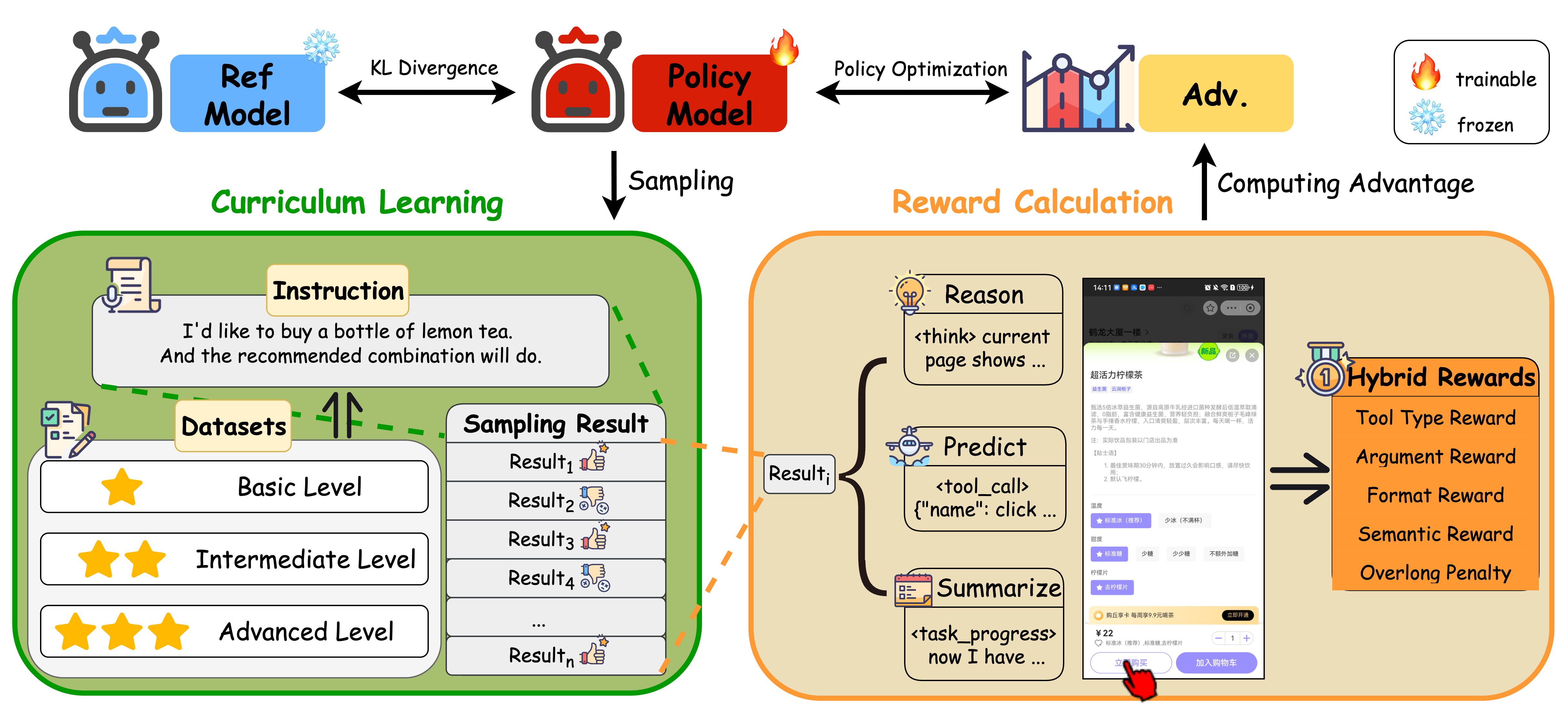}  
    \caption{Overview of the Training Framework. The policy model samples for a given task prompt from specific datasets. Then updating with fine-grained reward mechanism and GRPO objective. Iteration starts from basic level tasks to advanced level tasks.}  
    \label{fig:training_framework}
\end{figure}

\subsection{Curriculum-Guided GRPO Training}
We adopt Group Relative Policy Optimization (GRPO) as the foundational reinforcement learning algorithm in our framework. Unlike traditional methods such as PPO\cite{schulman2017proximal}, GRPO eliminates the need for a separate value function by leveraging group-wise advantage estimation, substantially reducing memory usage and computational overhead. This makes it particularly suitable for GUI-based training under limited hardware resources.

Our GRPO-based curriculum learning framework is illustrated in Figure~\ref{fig:training_framework}. Within this framework, the policy model undergoes gradient updates during training while the reference model remains frozen, with KL divergence regularization ensuring training stability. Training tasks are first partitioned according to operational complexity, where difficulty is defined by the number of interaction steps (i.e., trajectory length) required to complete an instruction. Tasks involving more steps are considered more challenging. This stratification yields three difficulty tiers:

\begin{enumerate}[label=\arabic*)]
\item \textbf{Stage 1: Basic level} - tasks requiring $\leq3$ interaction steps.
\item \textbf{Stage 2: Intermediate level} - tasks involving $4$ to $8$ steps, representing typical multi-step operational workflows.
\item \textbf{Stage 3: Advanced level} - tasks requiring $>8$ steps or involving visual understanding components such as VQA, information extraction, and element localization.
\end{enumerate}

Visual understanding tasks are also included in the Hard level due to their demand for fine-grained visual-semantic reasoning and contextual interpretation, which are essential capabilities for building general-purpose UI agents. Except for single-step tasks (path length = 1, typically representing low-level atomic actions), all levels conform to the high-level operation definitions established in industry standards. This progressive curriculum, advancing from basic UI manipulations to complex cross-modal cognition, facilitates human-like skill acquisition throughout training.

Each training batch follows a structured four-phase GRPO process: first, the policy model samples a group of $\{G\}$ candidate outputs $\{o_1, o_2, \ldots, o_G\}$ for a given task prompt; second, each output is evaluated using verifiable reward functions to produce scalar rewards $\{r_1, r_2, \ldots, r_G\}$; third, group-relative advantage values $\{\hat{A}_i\}$ are computed by normalizing the reward distribution within the group; finally, the policy $\pi_\theta$ is updated using the GRPO objective with the estimated advantages. The full training procedure is detailed in Algorithm~\ref{alg:igrpo}.

Notably, we follow the GRPO's original implement of advantage estimation. As shown in \ref{eq:adv}, the normalized advantage $\hat{A}_{i}$ is computed by the mean and standard deviation of the rewards respectively. This simplified approach eliminates intricate advantage computation procedures, significantly enhancing training efficiency while maintaining theoretical soundness.

\begin{equation}
    \hat{A_i} = \frac{r_i - \text{Mean}(\{r_1, r_2, \dots, r_G\})}{\text{Std}(\{r_1, r_2, \dots, r_G\})}
\label{eq:adv}
\end{equation}

\begin{algorithm}[!t]
\SetAlgoLined
\Input{initial policy model $\pi_{\theta_{\text{init}}}$; reward functions $r_\varphi$; task prompts $\mathcal{D}$}
\KwOut{$\pi_\theta$}
policy model $\pi_\theta \leftarrow \pi_{\theta_{\text{init}}}$\;

\For{each dataset $\mathcal{D}$ in curriculum order}{
    reference model $\pi_{\text{ref}} \leftarrow \pi_\theta$\;
    
    \For{$\text{step}=1,\dots,M$}{
        Sample a batch $\mathcal{D}_b$ from $\mathcal{D}$\;
        
        Update the old policy model $\pi_{\theta_{\text{old}}} \leftarrow \pi_\theta$\;
        
        Sample $G$ outputs $\{o_i\}_{i=1}^G \sim \pi_{\theta_{\text{old}}}(\cdot \mid q)$ for each question $q \in \mathcal{D}_b$\;
        
        Compute rewards $\{r_i\}_{i=1}^G$ for each sampled output $o_i$ by running $r_\varphi$\;
        
        Compute $\hat{A}_{i}$ for $o_i$ with ground truth $gt_i$ through group relative advantage estimation\;
        
        Update the policy model $\pi_\theta$ by maximizing the GRPO objective\;
    }
}
\caption{Curriculum GRPO for GUI Reasoning Model}
\label{alg:igrpo}
\end{algorithm}

\subsection{Fine-Grained Verified Reward Design}
To enable a more generalizable and capable UI agent, we design task-specific reward functions for both mobile operation and visual understanding tasks. Inspired by the rule-based reward paradigm introduced in DeepSeek-R1\cite{guo2025deepseek}—known for its simplicity and effectiveness—we adopt verified reward mechanisms that combine format validation and outcome assessment. This approach has recently expanded from LLMs to VLMs, as demonstrated in works such as VLM-R1~\cite{shen2025vlm}, and Visual-RFT~\cite{liu2025visual}. In our framework, most rewards are defined through verifiable rules, providing lightweight yet reliable signals to guide curriculum-based reinforcement learning.

\subsubsection{Mobile Operation Tasks}
For mobile operation tasks, the designed reward function is shown in Equation \ref{eq:operation_rewards}. We formulate CRAFT-GUI's action prediction as tool call for mobile operations, where distinct actions correspond to specific tools (e.g., click actions activate the click tool, swipe actions trigger the swipe tool). Accordingly, the operational rewards primarily derive from two aspects: 1) Correctness reward of tool selection $R_{tool}$, and 2) Accuracy reward of parameter specification in tool invocation $R_{args}$. A format reward $R_{format}$ is additionally incorporated to ensure structured output generation. $w$ represents the respective weight assigned to each reward.

\vspace{-9pt}
\begin{align}
R_{operation} =\ & w_{tool} * R_{tool} + w_{args} * R_{args} \notag \\
& + w_{f} * R_{format} + w_{l} * P_{length}
\label{eq:operation_rewards}
\end{align}

Each reward component is described in detail below to illustrate its role in guiding agent behavior.


\textbf{Tool Type Reward.}\hspace{1em}In mobile operation tasks, we follow Qwen2.5VL~\cite{bai2025qwen2}'s agent paradigm by formulating action spaces through agent tool call: \textit{click tool}, \textit{swipe tool}, \textit{type tool}, \textit{long press tool}, \textit{open app tool}, \textit{system button tool}, \textit{wait tool}, and \textit{terminate tool}. This toolkit achieves near-comprehensive coverage of terminal operation requirements. If the predicted tool $o_{tool}$ matches the ground truth tool ${gt}_{tool}$, it is assigned with a reward of 1, or 0 otherwise. As shown in Equation \ref{eq:tool_type}:

\begin{equation}
R_{tool} = 
\begin{cases} 
1 & \text{if } o_{tool} \text{ matches } gt_{tool}, \\
0 & \text{else.}
\end{cases}
\label{eq:tool_type}
\end{equation}

\textbf{Tool Arguments Reward.}\hspace{1em}Different tool invocations are associated with distinct parameter types:
\begin{enumerate}[label=\arabic*)]
    \item For \textit{click tool} and \textit{long press tool}, the parameter represents a point on screen with coordinate $(x, y)$, where x denotes pixels from the left edge and y indicates pixels from the top edge. As the ground truth is also a coordinate point, we adopt a specific verified rule. A prediction is rewarded if it falls within the same widget with the ground truth.
    \item For \textit{swipe tool}, parameters consist of two points: the start point $(x_1, y_1)$ and end point $(x_2, y_2)$. A prediction is considered valid if the angle between the predicted swipe vector and the ground-truth vector is within $30^\circ$.
    \item The \textit{type tool} and \textit{open app tool} require textual parameters. In these cases, we employ semantic similarity assessment combined with character-level matching to evaluate alignment with the ground truth text. 
    \item \textit{System button} tool takes one of several enumerated commands (e.g., Back, Home, Menu, Enter). Validation is performed via strict string matching.
    \item \textit{Wait} and \textit{terminate} tools require no arguments; thus, no reward is computed for these cases.
\end{enumerate}
All these rewards can be formulated as \ref{eq:paras}:
\begin{equation}
R_{paras} = 
\begin{cases} 
1 & \text{if } o_{paras} \text{ matches } gt_{paras}, \\
0 & \text{else.}
\end{cases}
\label{eq:paras}
\end{equation}

\textbf{Format reward.}\hspace{1em}
We design task-specific response templates to guide the model in producing structured reasoning processes prior to final outputs, following paradigms such as ReAct and chain-of-thought. To this end, we introduce thinking tokens, which enhance the model's logical reasoning capabilities and improve both stability and performance on complex tasks. All responses—across both mobile operation and visual understanding tasks—are formatted using HTML-style tags for consistency and interoperability.

For operation tasks, the response structure comprises three key components:
\begin{enumerate}[label=\arabic*)]
\item The $<think>$ tag explicitly documents the model's reasoning trajectory, enhancing output interpretability through transparent cognitive process visualization.
\item The $<tool\_call>$ tag standardizes action execution with JSON-formatted tool invocations, ensuring structured parameter specification and operational consistency.
\item The $<task\_progress>$ tag facilitates task state summarization and historical context preservation, functioning as an essential memory mechanism for multi-turn interaction.
\end{enumerate}

For a more detailed description of the format rewards for both operation and visual understanding tasks, please refer to the supplementary materials.
We formulate the format reward signal in Equation~\ref{eq:format_reward} as follows:
\begin{equation}
R_{format} = 
\begin{cases} 
1 & \text{if } o_i \text{ matches response pattern}, \\
0 & \text{else.}
\end{cases}
\label{eq:format_reward}
\end{equation}

\textbf{Overlong Response Penalty}\hspace{1em}During GRPO training we observe a progressive increase in response length, primarily driven by expanding thinking token sequences. If left unregulated, this phenomenon risks triggering explosive thinking token growth that may ultimately lead to model performance collapse. Inspired by DAPO~\cite{yu2025dapo}'s optimization principles, we implement an adaptive length constraint mechanism. This approach applies graduated penalties to responses exceeding predetermined length thresholds, effectively suppressing overgeneration tendencies while preserving essential reasoning components. As shown in Equation \ref{eq:length_penalty}, $L_{\max}$ is the max output length, and $L_{cache}$ is the buffer length for soft penalty.

\begin{equation}
P_{\text{length}} = 
\left\{
\begin{array}{ll}
0, & \! |o_{i}| \leq L_{\max} - L_{\text{cache}} \\
\frac{(L_{\max} - L_{\text{cache}}) - |o_{i}|}{L_{\text{cache}}}, & \! L_{\max} - L_{\text{cache}} < |o_{i}| \leq L_{\max} \\
-1, & \! L_{\max} < |o_{i}|
\end{array}
\right.
\label{eq:length_penalty}
\end{equation}

\subsubsection{Visual Understanding Tasks}\hspace{1em}
For visual understanding tasks covering visual question answering, information extraction, and element localization, we design a dedicated reward function as shown in Equation~\ref{eq:understanding_rewards}. These tasks enhance the agent’s ability to perceive and semantically interpret screen content. The semantic reward term $R_{sem}$ evaluates the alignment between model responses and annotated ground truths. Due to the complexity of natural language, $R_{sem}$ is based on a LLM-as-judged approach. A detailed implementation is provided in the supplementary material.

\begin{equation}
    R_{understanding} = w_{s} * R_{sem} + w_{f} * R_{format} + w_{l} * P_{length},
\label{eq:understanding_rewards}
\end{equation}

Specifically, we employ a format reward $R_{format}$ to enforce consistent output structures and incorporate an adaptive length penalty $P_{length}$ that progressively penalizes responses exceeding predefined length limits. Each reward component is weighted by its respective coefficient $w$.

\section{Experiments}
\subsection{Implement Details}
\textbf{Training and inference.}\hspace{1em}During our multi-stage curriculum reinforcement learning experiments, we evaluated performance scaling across varying model sizes by employing Qwen2.5VL with 7B, 32B parameters as base models. 
All experiments are conducted using computational resources equivalent to 96 NVIDIA A100 GPUs (80GB).
In terms of the consistency of the evaluation protocol, we maintain identical mobile operation and visual understanding prompts between the training and inference phases. Model capabilities are assessed through: 1) zero-shot generalization on unseen task distributions; 2) pass@1 success rate under strict single-attempt execution constraints

\textbf{Evaluation benchmarks and metrics.}\hspace{1em}We established an online evaluation benchmark to comprehensively assess model capabilities in real-world operational environments, focusing on three critical dimensions: page comprehension accuracy, contextual reasoning effectiveness, and integrated decision-making proficiency. To validate the methodological efficacy proposed in this work, parallel evaluations were conducted on open-source benchmark suites including Android Control\cite{li2024effects} and AndroidWorld\cite{rawles2024androidworld}. The primary evaluation metric employed is task success rate (SR), calculated as Equation \ref{eq:sr}:
\begin{equation}
    SR = \frac{N_{\text{success}}}{N_{\text{total}}} \times 100\%
\label{eq:sr}
\end{equation}

\subsection{Experimental Results}
To validate the effectiveness of our proposed methodology – comprising multi-stage curriculum reinforcement learning and joint training of operation-understanding tasks – we conducted systematic evaluations across both in-house and open-source datasets.

\begin{table}[h]
\centering
\caption{Comparison of Agent Models on AndroidControl-Low, AndroidControl-High and AndroidWorld; * Claude refers Claude-computer-use.}
\label{tab:android_control}
\footnotesize
\setlength{\tabcolsep}{4pt}
\begin{tabular}{l c c c c c c c}
\toprule
\multirow{2}{*}{\textbf{Agent Models}} 
& \multicolumn{3}{c}{\textbf{AndroidControl-Low}} 
& \multicolumn{3}{c}{\textbf{AndroidControl-High}}
& \textbf{AndroidWorld} \\
\cmidrule(lr){2-4} \cmidrule(lr){5-7} \cmidrule(lr){8-8}
& \textbf{Type} & \textbf{Grounding} & \textbf{SR} 
& \textbf{Type} & \textbf{Grounding} & \textbf{SR} 
& \textbf{SR} \\
\midrule
Claude$^*$ & 74.3\% & 0.0\% & 19.4\% & 63.7\% & 0.0\% & 12.5\% & 27.9\% \\
GPT-4o & 74.3\% & 0.0\% & 19.4\% & 66.3\% & 0.0\% & 20.8\% & 34.5\% \\
SeeClick & 93.0\% & 73.4\% & 75.0\% & 82.9\% & 62.9\% & 59.1\% & -- \\
InternVL-2-4B & 90.9\% & 84.1\% & 80.1\% & 84.1\% & 72.7\% & 66.7\% & -- \\
Qwen2-VL-7B & 91.9\% & 86.5\% & 82.6\% & 83.8\% & 77.7\% & 69.7\% & -- \\
Aria-UI & -- & 87.7\% & 67.3\% & -- & 43.2\% & 10.2\% & -- \\
OS-Atlas-7B & 93.6\% & 88.0\% & 85.2\% & 85.2\% & 78.5\% & 71.2\% & -- \\
Aguvis-72B & -- & -- & 84.4\% & -- & -- & 66.4\% & 26.1\% \\
UI-TARS-72B & 98.1\% & 89.9\% & 91.3\% & 85.2\% & 81.5\% & 74.7\% & 46.6\% \\
Qwen2.5-VL-32B & -- & -- & 93.3\% & -- & -- & 69.6\% & 22.0\% \\
Qwen2.5-VL-72B & -- & -- & 93.7\% & -- & -- & 67.4\% & 35.0\% \\
\midrule
CRAFT-GUI-32B-stage1 & 98.0\% & 92.4\% & 91.3\% & 88.7\% & 81.9\% & 75.6\% & 41.4\% \\
CRAFT-GUI-32B-stage2 & 98.8\% & 93.1\% & 92.2\% & 90.8\% & 84.4\% & 78.7\% & 44.8\% \\
CRAFT-GUI-32B-stage3 & \textbf{98.9\%} & \textbf{93.6\%} & 92.7\% & \textbf{91.0\%} & \textbf{85.8\%} & \textbf{80.3\%} & \textbf{51.7\%} \\
\bottomrule
\end{tabular}
\end{table}

\subsubsection{Open-source Datasets Training and Evaluation}
To validate the feasibility of our proposed methodology, we conducted training on public datasets and performed corresponding evaluations on designated benchmarks. Specifically, we selected the offline evaluation platforms - Android Control - along with the online evaluation platform AndroidWorld for comprehensive testing.

For the Android Control training set, we implemented curriculum reinforcement learning training using the Qwen2.5-VL-32B as our base model. The training configuration strictly followed our proposed methodology. Task trajectories are stratified into basic, intermediate, and advanced difficulty levels based on task path complexity metrics. And the training progress is executed with our hybrid reward mechanism. During evaluation, we employ the complete Android Control test set as our benchmark, maintaining strict separation between training and evaluation data splits. For Android Control, we report two settings (low and high). Meanwhile we use the same weights to evaluate on AndroidWorld. As shown in Table \ref{tab:android_control}, the performance of CRAFT-GUI-32B demonstrates the effectiveness of our proposed method.

\subsubsection{In-house Datasets Training and Evaluation}

We developed a proprietary training dataset comprising 80K samples across six major categories of mobile applications: food delivery, in-store dining, medical services, financial services, insurance services, and gaming apps. All data are collected from multiple device platforms (iPhone, Android, iPad) to ensure cross-platform generalization. This multi-domain corpus is organized into three difficulty levels to support phased curriculum learning: Basic level (21K operation samples, $\leq$3 steps), Intermediate level (28K operation samples, 4-8 steps), and Advanced level (8K operation samples and 23K understanding samples, $>$8 steps or involving visual reasoning tasks).

Our method significantly outperforms industry baselines including Claude-3.7-Sonnet~\cite{anthropic2024cuda} and GPT-4.1~\cite{openai2025gpt41} across all application domains. CRAFT-GUI-32B achieves 75.7\% average success rate, representing 10.3\% improvement over the best baseline (detailed results in Appendix Table~\ref{tab:sr_comparison_model}). Progressive curriculum learning yields consistent gains: our 32B model improves from 62.6\% (Stage 1) to 75.7\% (Stage 3), while the 7B variant progresses from 43.9\% to 64.1\% (see Appendix Table~\ref{tab:internal_sr_7B} for stage-wise breakdown). Additional understanding task results are provided in Appendix Table~\ref{tab:acc_comparison_model}.

\subsection{Ablation Study}

\begin{table}[htbp!]
    \centering
    \caption{SR of Different Training Method}
    \label{tab:sr_comparison_training_method}
    \footnotesize
    \setlength{\tabcolsep}{4pt}
    \begin{tabular}{cccccccc}
        \toprule
        \multirow{2}*{\textbf{Training Method}} & 
        \multicolumn{6}{c}{\textbf{Mobile Application Categories}} & 
        \multirow{2}*{\textbf{Average}} \\
        \cmidrule(lr){2-7}
        & \textbf{Food-delivery} & \textbf{Dine-in} & \textbf{Medical} & \textbf{Finance} & \textbf{Insurance} & \textbf{Gaming} & \\
        \midrule    
        SFT & 52.9\% & 48.6\% & 62.7\% & 58.8\% & 48.5\% & 91.5\% & 60.8\% \\
        Vanilla GRPO & 60.8\% & 69.2\% & 70.6\% & 75.5\% & \textbf{62.4\%} & 92.6\% & 71.9\% \\
        Curriculum GRPO & \ \textbf{76.5\%} & \textbf{73.8\%} & \textbf{73.5\%} & \textbf{77.5\%} & 60.4\% & \textbf{92.7\%} & \textbf{75.7\%} \\
        \bottomrule
    \end{tabular}
\end{table}

\begin{table}[htbp!]
    \centering
    \caption{SR of Different Data Mixture}
    \label{tab:comparison_data_mix}
    \footnotesize
    \setlength{\tabcolsep}{4pt}
    \begin{tabular}{cccccccc}
        \toprule
        \multirow{2}*{\textbf{Training Data}} & 
        \multicolumn{6}{c}{\textbf{Mobile Application Categories}} & 
        \multirow{2}*{\textbf{Average}} \\
        \cmidrule(lr){2-7}
        & \textbf{Food-delivery} & \textbf{Dine-in} & \textbf{Medical} & \textbf{Finance} & \textbf{Insurance} & \textbf{Gaming} & \\
        \midrule
        Operation Only & 67.6\% & 72.0\% & 73.3\% & 74.3\% & 59.4\% & 91.5\% & 73.2\% \\
        Operation + Understanding & \textbf{76.5\%} & \textbf{73.8\%} & \textbf{73.5\%} & \textbf{77.5\%} & \textbf{60.4\%} & \textbf{92.7\%} & \textbf{75.7\%} \\
        \bottomrule
    \end{tabular}
\end{table}

We conduct ablation experiments on the 32B-scale model to systematically demonstrate the effects of different ablated components. The ablated models are trained on our proprietary internal dataset and thoroughly evaluated on the online benchmark comprising six categories of mobile applications. The evaluation results are presented in tabular format for clear visualization. The ablation study outcomes conclusively validate the effectiveness of our proposed methodology.

\textbf{The Effectiveness of Curricumlum RL Strategy}\hspace{1em}In the experiments of training strategies, we utilized identical raw datasets. For the three strategy, each sample's system prompt, user prompt, and assistant prompt remain strictly consistent across training configurations. Moreover, the training datasets all incorporate operation-type and understanding-type data. As shown in the Table \ref{tab:sr_comparison_training_method}, the curriculum reinforcement learning algorithm achieves the highest average operation success rate of 75.7\% across six categories of mobile applications. The traditional reinforcement learning implementation demonstrates an 11.1\% improvement over the supervised fine-tuning (SFT) baseline, while the curriculum reinforcement learning approach yield a 14.9\% enhancement compared to SFT. Furthermore, applying curriculum learning strategies to the reinforcement learning framework provides an additional 3.8\% performance gain over the standard reinforcement learning implementation.

\textbf{The Effectiveness of Training Data Mixture}\hspace{1em}Based on the curriculum reinforcement learning training strategy, we investigate the necessity of mixing operation tasks and understanding tasks. As shown in Table \ref{tab:comparison_data_mix}, we conduct two sets of experiments under identical curriculum reward training strategy settings: one group containing only operation-type data, while the other incorporates complex and challenging understanding-type data in the third phase. The results demonstrate that with the inclusion of understanding-type data, the model exhibits enhanced comprehension and mastery of operation-type tasks, achieving a 2.5\% improvement in execution success rate compared to using single-type data.  

\section{Conclusion}
In this work, we present CRAFT-GUI, a GUI reasoning model developed through a novel multi-stage reinforcement learning training framework. Owing to the joint training of operation tasks and understanding tasks, CRAFT-GUI thereby advances towards a more versatile GUI-Agent system that autonomously reasons and interacts with mobile devices to complete user instructions. The carefully engineered reward and penalty mechanisms ensure stable model training while maintaining robust performance metrics in evaluations. Additional training details and case studies can be found in the Appendix section. 
In future work, we plan to extend CRAFT-GUI to computer tasks and introduce trial-and-error with rollback, enabling more general-purpose intelligent agents.

\bibliography{reference}
\bibliographystyle{iclr2026_conference}

\appendix
\section{Additional Training details}

Understanding tasks adopt a dual-phase structure, as shown in Figure\ref{fig:understanding_template}:
\begin{enumerate}[label=\arabic*)]
\item The $<think>$ tag enables multi-step semantic reasoning, capturing hierarchical understanding of visual-semantic content.
\item The $<answer>$ tag delivers final responses with precise information encapsulation, strictly following predefined knowledge representation formats.
\end{enumerate}

\begin{figure}[htbp]  
    \centering  
    \includegraphics[width=0.5\textwidth]{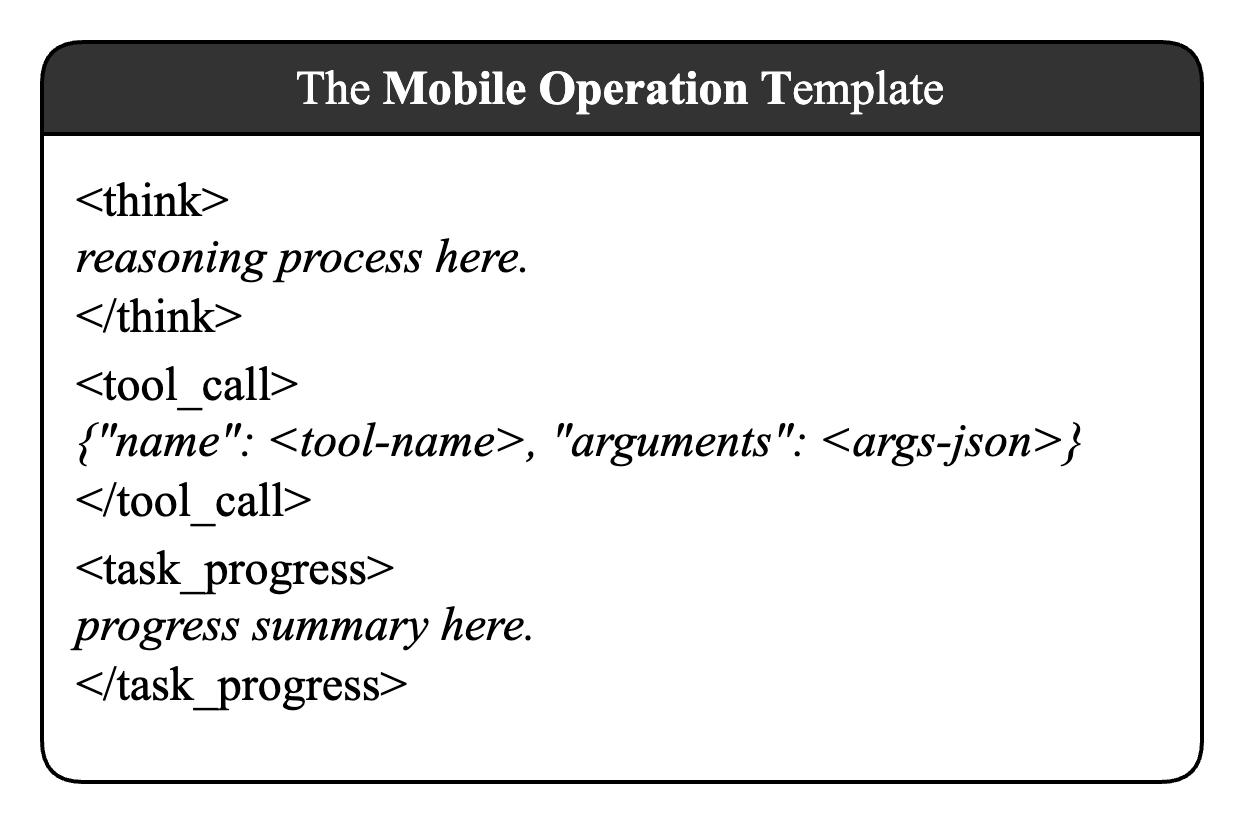}  
    \caption{The Mobile Operation Template.}
    \label{fig:operation_template}
\end{figure}

\begin{figure}[htbp]  
    \centering  
    \includegraphics[width=0.4\textwidth]{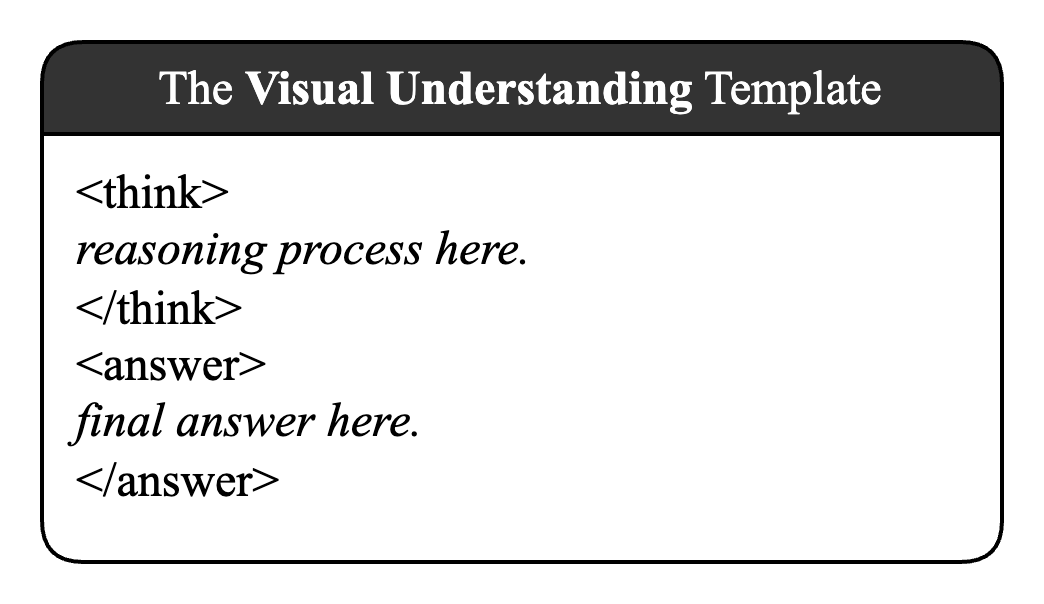}  
    \caption{The Visual Understanding Template.}
    \label{fig:understanding_template}
\end{figure}

\begin{figure}[htbp]  
    \centering  
    \includegraphics[width=0.8\textwidth]{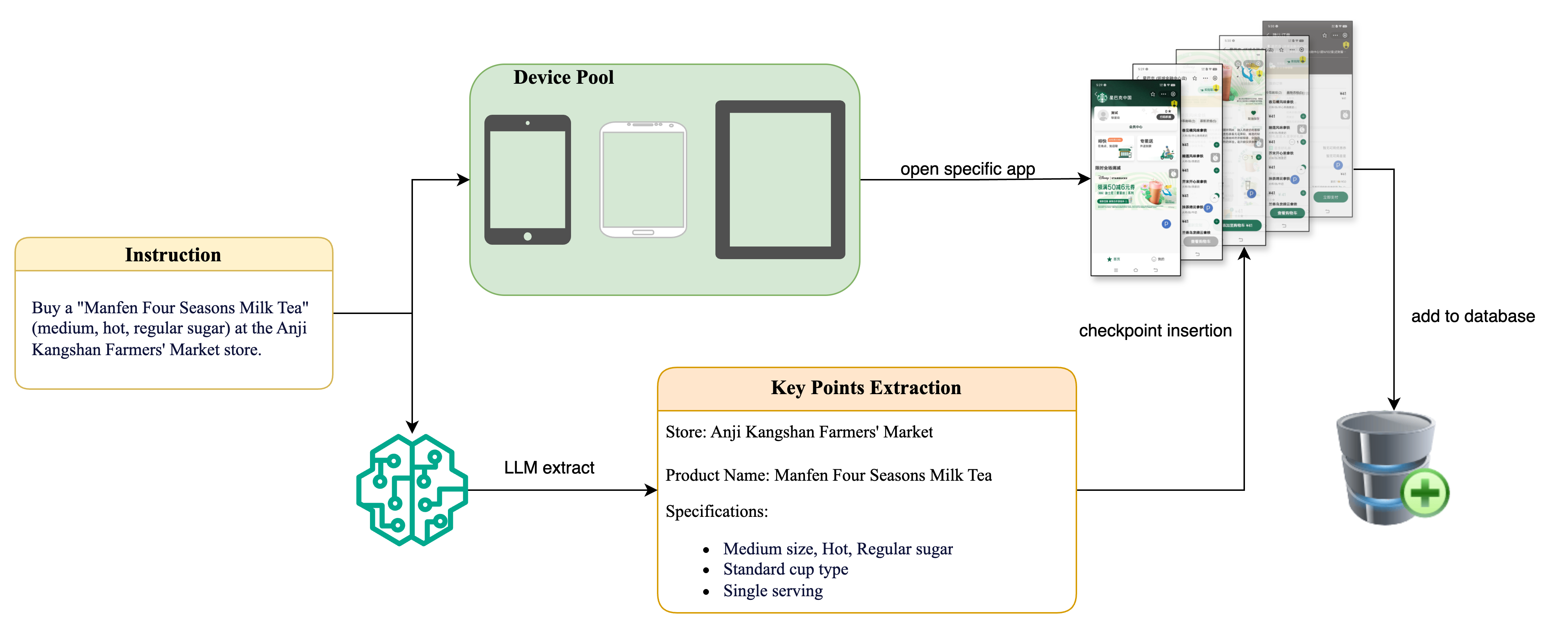}  
    \caption{Illustration of Our Online Benchmark building pipeline.}  
    \label{fig:online_bench_build}  
\end{figure}

Figure \ref{fig:online_bench_build} illustrates the workflow used to construct and run an evaluation episode. \textcolor{blue}{For every task, we (i) reconstruct the corresponding UI flow on an emulator, (ii) tag critical intent checkpoints such as item selection, quantity confirmation, or payment submission, and (iii) introduce stochastic branches (e.g., pop-ups, alternative menus) together with latency or layout noise.} During execution the evaluator records success at each checkpoint, enabling fine-grained and realistic performance measurement.

\textbf{Semantic Consistency Reward}\hspace{1em}In operation tasks involving textual parameters and visual understanding tasks, semantic consistency reward is typically involved. This reward quantifies the textual similarity between the model's output $o_{sem}$ and the ground truth $gt_{sem}$, where higher similarity corresponds to greater reward values. Traditional lexical-based metrics (e.g., TF-IDF, Rouge-L) prove insufficient to capture nuanced semantic equivalence due to linguistic diversity, particularly in handling synonymy, paraphrasing, and contextual awareness.

To address this limitation, we remotely deploy small language reward model for semantic evaluation. This model implements a comprehensive multi-dimensional assessment framework, considering:
\begin{enumerate}[label=\arabic*)]
    \item Entity consistency - verification of critical information alignment
    \item Content alignment - semantic equivalence of descriptive elements
    \item Contextual emphasis - preservation of key focus points
    \item Paraphrase robustness - tolerance to syntactical variations
\end{enumerate}

This design effectively bridges the gap between rigid pattern matching and flexible semantic understanding in reward assignment. The scoring process can be formulated as Equation \ref{eq:sem_reward}, where $\mathcal{S}$ means the semantic similarity between output $o_{sem}$ and the ground truth $gt_{sem}$.

\begin{equation}
R_{sem} \in \left\{ 0,\, 0.2,\, 0.4,\, \dots,\, 1 \right\} \quad \text{depends on } \mathcal{S}(o_{sem}, gt_{sem})
\label{eq:sem_reward}
\end{equation}

\section{Online Benchmark for Operation Task}
\begin{figure}[htbp]  
    \centering  
    \includegraphics[width=0.98\textwidth]{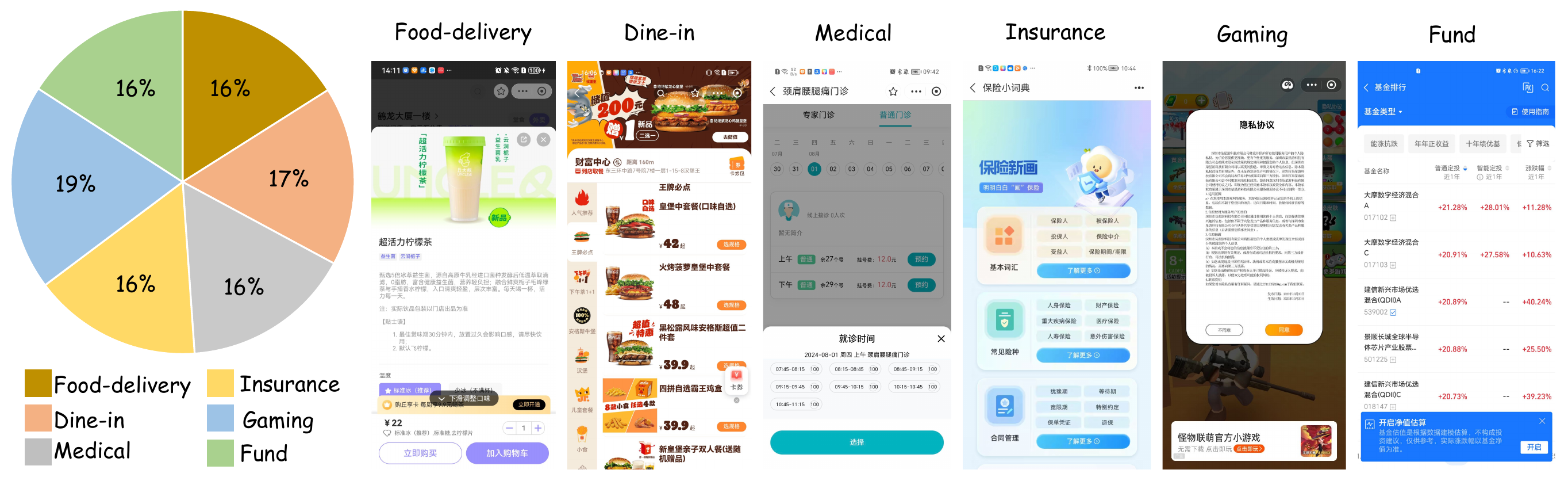}  
    \caption{Domain-specific Distribution of Our Online Benchmark.}  
    \label{fig:online_bench}  
\end{figure}
Interactive applications deployed in the real world—such as food delivery, medical appointment scheduling, navigation, and ticketing—exhibit highly variable user flows. To be useful, an agent must adapt to individual habits and dynamically changing requirements.

We constructed an online benchmark encompassing six categories of daily-use mobile applications: food delivery, in-store dining, medical services, financial services, insurance services, and gaming applications. Each category incorporates a diverse selection of mobile apps reflecting real-world operational complexity. The benchmark composition follows domain-specific distribution as Figure \ref{fig:online_bench}:

The benchmark's tasks are statistically derived from real-human interaction logs, preserving authentic operational frequency across application domains. Each category contains 120-200 unique test cases with configurable difficulty levels, supporting both functional verification and cognitive capability assessment.

\begin{table}[t]
    \centering
    \caption{SR (Success Rate, Pass@1) on Internal Online Operation Benchmark}
    \label{tab:sr_comparison_model}
    \footnotesize
    \setlength{\tabcolsep}{4pt}
    \begin{tabular}{lccccccr}
        \toprule
        \multirow{2}*{\textbf{Model}} & 
        \multicolumn{6}{c}{\textbf{Mobile Application Categories}} & 
        \multirow{2}*{\textbf{Average}} \\
        \cmidrule(lr){2-7}
        & \textbf{Food-delivery} & \textbf{Dine-in} & \textbf{Medical} & \textbf{Finance} & \textbf{Insurance} & \textbf{Gaming} & \\
        \midrule
         Claude-3.7-Sonnet & 31.7\% & 28.3\% & 44.4\% & 24.2\% & 25.6\% & 61.7\% & 35.9\% \\
         \midrule
         GPT-4.1 & 19.5\% & 18.8\% & 32.4\% & 21.3\% & 18.0\% & 56.2\% & 27.7\% \\
         \midrule  
         UI-TARS-72B & 65.4\% & 57.2\% & 67.5\% & 61.8\% & 55.1\% & 85.3\% & 65.4\% \\
         \midrule
         Qwen2.5-VL-72B & 65.7\% & 52.3\% & 67.6\% & 57.4\% & 56.9\% & 88.6\% & 64.8\% \\
         \midrule
         CRAFT-GUI-7B & 43.9\% & 48.4\% & 74.5\% & 67.4\% & 54.2\% & 96.4\% & 64.1\% \\
         CRAFT-GUI-32B & \textbf{76.5\%} & \textbf{73.8\%} & \textbf{73.5\%} & \textbf{77.5\%} & \textbf{60.4\%} & 92.7\% & \textbf{75.7\%} \\
        \bottomrule
    \end{tabular}
\end{table}

\section{More Samples}

\textbf{CRAFT-GUI-7B}\hspace{1em}Table \ref{tab:internal_sr_7B} presents the evaluation outcomes of the 7B model across distinct curriculum reinforcement learning stages on the online benchmark. The results demonstrate progressive success rate enhancements across all mobile application categories as training progresses through sequential phases.
\begin{table}[htbp]
    \centering
    \caption{CRAFT-GUI-7B Performance (SR) Across Mobile Application Categories by RL Period}
    \label{tab:internal_sr_7B}
    \footnotesize
    \setlength{\tabcolsep}{4pt}
    \begin{tabular}{lccccccr}
        \toprule
        \multirow{2}*{RL Period} & 
        \multicolumn{6}{c}{Mobile Application Categories} & 
        \multirow{2}*{Average} \\
        \cmidrule(lr){2-7}
        & Food-delivery & Dine-in & Medical & Finance & Insurance & Gaming & \\
        \midrule
         Stage 1 & 24.3\% & 17.2\% & 38.4\% & 50.2\% & 40.8\% & 92.7\% & 43.9\% \\
         Stage 2 & 26.3\% & 29.2\% & 58.5\% & 57.8\% & 47.8\% & 95.0\% & 52.4\% \\
         Stage 3 & \textbf{43.9\%} & \textbf{48.4\%} & \textbf{74.5\%} & \textbf{67.4\%} & \textbf{54.2\%} & \textbf{96.4\%} & \textbf{64.1\%} \\
        \bottomrule
    \end{tabular}
\end{table}

\begin{table}[htbp]
    \centering
    \caption{CRAFT-GUI-32B Performance (SR) Across Mobile Application Categories by RL Period}
    \label{tab:internal_sr_32B}
    \footnotesize
    \setlength{\tabcolsep}{4pt}
    \begin{tabular}{lccccccr}
        \toprule
        \multirow{2}*{RL Period} & 
        \multicolumn{6}{c}{Mobile Application Categories} & 
        \multirow{2}*{Average} \\
        \cmidrule(lr){2-7}
        & Food-delivery & Dine-in & Medical & Finance & Insurance & Gaming & \\
        \midrule
        Stage 1 & 51.0\% & 55.1\% & 63.7\% & 61.8\% & 51.5\% & 92.7\% & 62.6\% \\
        Stage 2 & 59.8\% & 72.9\% & 66.7\% & 68.6\% & 58.4\% & \textbf{93.5\%} & 69.9\% \\
        Stage 3 & \textbf{76.5\%} & \textbf{73.8\%} & \textbf{73.5\%} & \textbf{77.5\%} & \textbf{60.4\%} & 92.7\% & \textbf{75.7\%} \\
        \bottomrule
    \end{tabular}
\end{table}

\textbf{CRAFT-GUI-32B}\hspace{1em}Table \ref{tab:internal_sr_32B} details the evaluation results of the 32B model under identical training framework configurations on the online benchmark. The expanded parameter space of the base model yielded substantial performance gains, with the 32B variant achieving 11.6\% higher average success rate compared to the 7B model during final curriculum phase evaluation.  

Progressive performance escalation is observed throughout the three-stage curriculum learning process: stage 1 (62.6\% SR), stage 2 (69.9\% SR), and final stage (75.7\% SR), demonstrating large relative improvement from initial to final training completion.

\textbf{Understanding metrics}
For visual understanding scenarios such as visual question answering, information extraction and element localization, we construct respective benchmarks covering six kinds of mobile applications (Food delivery, et al.). The evaluation metric employed is accuracy (Acc), calculated as Equation \ref{eq:acc}:
\begin{equation}
    Acc = \frac{N_{\text{correct}}}{N_{\text{total}}} \times 100\%
\label{eq:acc}
\end{equation}

\begin{table}[htbp]
    \centering
    \caption{Acc on Visual Understanding Benchmark}
    \label{tab:acc_comparison_model}
    \footnotesize
    \setlength{\tabcolsep}{4pt}
    {\begin{tabular}{lc}
        \toprule
        \multirow{2}*{Model} & 
        \multicolumn{1}{c}{Average} \\
        \cmidrule(lr){2-2}
         & VQA \& Info Extraction \& Localization \\
        \midrule
         Claude-3.7-Sonnet & 90.4\%\\
         \midrule
         GPT-4.1 &  91.8\%\\
         \midrule  
         CRAFT-GUI-7B & 80.5\%\\
         CRAFT-GUI-32B & \textbf{94.0\%}\\
        \bottomrule
    \end{tabular}
    }
\end{table}

\textbf{The Effectiveness of Overlong Response Penalty}\hspace{1em}Since the training process incorporates understanding-related tasks, the think tokens exhibit uncontrollable growth during training progression. To address this, we conduct ablation experiments to investigate the impact of applying an overlong response penalty on model performance. In this experimental group, while maintaining identical configurations of training data, three-phase curriculum reinforcement learning strategy, and multi-task data mixing protocol with the baseline, we solely introduce the length penalty constraint. The ablation study reveals that without length penalty regulation, the model progressively generates excessive nonsensical phrases and enters infinite reasoning loops until exceeding the maximum sequence length threshold, ultimately leading to catastrophic performance collapse. In contrast, with the application of response length penalty, the training process remains stable and achieves the anticipated performance benchmarks. Details on the online benchmark can be found in Table \ref{tab:comparison_overlong}.

\begin{table}[htbp!]
    \centering
    \caption{SR of Different Penalty Strategy}
    \label{tab:comparison_overlong}
    \footnotesize
    \setlength{\tabcolsep}{4pt}
    \begin{tabular}{cccccccc}
        \toprule
        \multirow{2}*{Training Data} & 
        \multicolumn{6}{c}{Mobile Application Categories} & 
        \multirow{2}*{Average} \\
        \cmidrule(lr){2-7}
        & Food-delivery & Dine-in & Medical & Finance & Insurance & Gaming & \\
        \midrule
        w/o Overlong Penalty & 59.6\% & 61.9\% & 64.5\% & 64.3\% & 48.4\% & 82.6\% & 63.5\% \\
        w/ Overlong Penalty & \textbf{76.5\%} & \textbf{73.8\%} & \textbf{73.5\%} & \textbf{77.5\%} & \textbf{60.4\%} & \textbf{92.7\%} & \textbf{75.7\%} \\
        \bottomrule
    \end{tabular}
\end{table}

\begin{table}[htbp]
    \centering
    \caption{Acc of Different Training Method}
    \label{tab:acc_comparison_training_method}
    \footnotesize
    \setlength{\tabcolsep}{4pt}
    \begin{tabular}{cc}
        \toprule
        \multirow{2}*{Model} & 
        \multicolumn{1}{c}{Average} \\
        \cmidrule(lr){2-2}
         & VQA \& Info Extraction \& Localization \\
         \midrule  
         SFT & 82.2\%\\
         Vanilla GRPO & 91.4\%\\
         Curriculum GRPO & \textbf{94.0\%}\\
        \bottomrule
    \end{tabular}
\end{table}

As shown in the Table \ref{tab:acc_comparison_training_method}, the curriculum RL also achieves the highest average accuracy of 94.0\% on our visual understanding benchmark across three types of task. The vanilla RL method get 9.2\% improvement over the SFT baseline. And curriculum RL get another 2.6\% enhancement over the standard RL.

\end{document}